\relax
\documentclass{article}
\usepackage{ijcai18}
\usepackage{times}
\usepackage{soul}
\usepackage[utf8]{inputenc}
\usepackage[small]{caption}
\usepackage[T1]{fontenc}
\usepackage{url}
\usepackage{graphicx} 
\usepackage{amsmath}
\usepackage{mathtools}
\usepackage{amssymb}
\usepackage{listings}
\usepackage[ruled]{algorithm2e}
\usepackage{booktabs}
\usepackage{subcaption}
\usepackage{float}
\usepackage{amsfonts}
\usepackage{booktabs}
\usepackage{siunitx}
\usepackage[table]{xcolor}
\graphicspath{{figs/}{../figures/}}
\newcommand{\graycell}{\cellcolor{gray!25}}

\usepackage{fancyhdr}
\begin{document}

\title{Objective-Reinforced Generative Adversarial Networks (ORGAN) for Sequence Generation Models}


\author{ Gabriel Guimaraes\textsuperscript{*,$\dagger$} \; Benjamin Sanchez-Lengeling\textsuperscript{*} \; Carlos Outeiral\textsuperscript{*,\textsection
} \\ { \bf \Large{Pedro Luis Cunha Farias}\textsuperscript{*} } \;  { \bf \Large{Alán Aspuru-Guzik}\textsuperscript{*,$\ddagger$} } \\ *\;Harvard University\\ $\dagger$\; Pagedraw\\ \textsection\; The University of Manchester\\ $\ddagger$\; Canadian Institute for Advanced Research (CIFAR) Senior Fellow}

\maketitle

\begin{abstract}

In sequence-based generative models, besides the generation of samples likely to have been drawn from a data distribution, it is often desirable to fine-tune the samples towards some domain-specific metrics.  This work proposes a method to guide the structure and quality of samples utilizing a combination of adversarial training and expert-based rewards with reinforcement learning. Building on SeqGAN, a sequence based Generative Adversarial Network (GAN) framework modeling the data generator as a stochastic policy in a reinforcement learning setting, we extend the training process to include domain-specific objectives additional to the discriminator reward. The mixture of both types of rewards can be controlled via a tune-able parameter. To improve training stability we utilize the Wasserstein distance as loss function for the discriminator. We demonstrate the effectiveness of this approach in two tasks: generation of molecules encoded as text sequences and musical melodies. The experimental results demonstrate the models can generate samples which maintain information originally learned from data, retain sample diversity, and show improvement in the desired metrics.


\end{abstract}

\makeatletter
\def\blfootnote{\gdef\@thefnmark{}\@footnotetext blah}
\makeatother

\section{Introduction}

Unsupervised generation of data is a dynamic area of machine learning and a very active research
frontier in areas ranging from language processing and music generation to materials and drug discovery. 

In any of these fields, it is often advantageous to guide the generative model towards some desirable characteristics, while ensuring that the samples resemble the initial distribution. In music generation, for example, it might be expected that pleasant melodic patterns prevail over more dissonant ones \cite{SequenceTutor}. In natural language processing, a given sentiment might be emphasized, maybe for producing movie reviews \cite{radford2017learning}. Finally, in materials discovery, the aim is often to optimize some properties for a particular application, for example in organic solar cells \cite{Hachmann2011}, OLEDs \cite{Gomez-Bombarelli2016a} or new drugs. 
The generation of discrete data using Recurrent Neural Networks (RNNs), in particular, Long Short-Term Memory cells \cite{LSTM} and maximum likelihood estimation has been shown to work well in practice. However, this often suffers from the so-called \textit{exposure bias}, and might lack some of the multi-scale structures or salient features of the data. Meanwhile Generative Adversarial Networks (GANs) \cite{Ian14}, an approach where a generative model competes against a
discriminate model, one trying to generate likely data while the other trying to distinguish false from real data. GANs have shown remarkable results at generation of data that imitates a data distribution, however they can suffer from several issues, among these mode-collapse \cite{Arjovsky2017}. Where the generator learns to produce samples with low variety.

Although GANs were not initially applicable to discrete data due to non-differentiability, approaches such as SeqGAN \cite{SeqGAN}, MaliGAN \cite{Che2017} and BGAN \cite{Hjelm2017} have arisen to deal with this issue.

Furthermore methods from Reinforcement Learning (RL) have shown great success at solving problems where continuous feedback from an environment is needed \cite{Hjelm2017}. 

In this paper, we introduce a novel approach to optimize the properties of a distribution of sequences, increase the diversity of the samples while maintaining the likeliness of the data distribution. In our approach, the generator is trained to maximize a weighted average of two types of rewards: the \textit{objective}, domain-specific metrics, and the \textit{discriminator}, which is trained along with the generator in an adversarial fashion. While the objective component of the reward function ensures that the model selects for traits that maximize the specified heuristic, the discriminator incentives the samples to stay within boundaries of the initial data distribution. Diversity is additionally promoted by reducing rewards of non-unique and less diverse sequences.

In order to implement the above idea, we build on SeqGAN, a recent work that successfully combines GANs and RL to apply the GAN framework to sequential data \cite{SeqGAN} and extend it towards domain-specific rewards. To increase the stability of the adversarial training, we test Wasserstein-GANs \cite{Arjovsky2017a} in this framework.

We test our model in the context of molecular and music generation, optimizing several domain-specific metrics. Our results show that ORGAN is able to tune the quality and structure of samples. We compare our results with the maximum likelihood estimation (MLE), SeqGAN and a RL approach.




\section{Related work}

Previous work has relied on specific modifications of the objective function to reach the desired properties. For example, \cite{SequenceTutor} introduce penalties to unrealistic sequences, in absence of which RL can easily get stuck around local maxima which can be very far from the global maximum reward. Related applications by \cite{ranzato2015sequence} and \cite{li2016deep} apply reinforcement learning to sequence generation in a NLP setting.

In the last two years, many methodologies have been proposed for \textit{de novo} molecular generation. \cite{ertl2017silico} and \cite{segler2017generating} trained recurrent neural networks to generate drug-like molecules. \cite{HIPSVAE} employed a variational autoencoder to build a latent, continuous space where property optimization can be made through surrogate optimization. Finally, \cite{kadurin2017drugan} presented a GAN model for drug generation. Additionally, the approach presented in this paper has recently been applied to molecular design \cite{Sanchez-Lengeling2017}.

In the field of music generation, \cite{lee2017seqgan} built a SeqGAN model employing an efficient representation of multi-channel MIDI to generate polyphonic music. \cite{chen2017learning} presented Fusion GAN, a dual-learning GAN model that can fuse two data distributions. \cite{jaques2017tuning} employ deep Q-learning with a cross-entropy reward to optimize the quality of melodies generated from an RNN.

In adversarial training, \cite{Pfau2016} recontextualizes GANs in the actor-critic setting. This connection is also explored with the Wasserstein-1 distance in WGANs \cite{Arjovsky2017a}.  Minibatch discrimination and feature mapping were used to promote diversity in GANs \cite{Ian16}.
Another approach to avoid mode collapse was shown with Unrolled GANs \cite{Metz2016}.
Issues and convergence of GANs has been studied in \cite{Mescheder2017}.

\section{Background}

In this section, we elaborate on the GAN and RL setting based on SeqGAN \cite{SeqGAN}

$G_\theta$ is a generator parametrized by $\theta$, that is trained to produce high-quality sequences $Y_{1:T} = (y_1, ..., y_T)$ of length $T$ and a discriminator model $D_\phi$ parametrized by $\phi$, trained to classify real and generated sequences. $G_\theta$ is trained to deceive $D_\phi$, and  $D_\phi$ to classify correctly. Both models are trained in alternation, following a minimax game:

\begin{equation}
\min_{\phi} \mathbb{E}_{Y \sim p_{\text{data}}(Y)} \left[\log D(Y) \right] + \mathbb{E}_{Y\sim p_{G_\theta}(Y)} \left[\log (1 - D(Y)) \right]
\end{equation}

For discrete data, the sampling process is not differentiable. However, $G_\theta$ can be trained as an agent in a reinforcement learning context using the REINFORCE algorithm \cite{REINFORCE}. Let $R(Y_{1:T})$ be the reward function defined for full length sequences. Given an incomplete sequence $Y_{1:t}$, also to be referred to as state $s_t$, $G_\theta$ must produce an action $a$, along with the next token $y_{t+1}$. 

The agent's stochastic policy is given by $G_\theta(y_t | Y_{1:t-1})$ and we wish to maximize its expected long term reward

\begin{equation}
J(\theta) = E[R(Y_{1:T}) | s_0, \theta] = \sum_{y_1 \in Y} G_\theta(y_1 | s_0) \cdot Q(s_0, y_1)
\end{equation}

where $s_0$ is a fixed initial state. $Q(s, a)$ is the action-value function that represents the expected reward at state $s$ of taking action $a$ and following our current policy $G_\theta$ to complete the rest of the sequence. For any full sequence $Y_{1:T}$, we have $Q(s = Y_{1:T-1}, a = y_T) = R(Y_{1:T})$ but we also wish to calculate $Q$ for partial sequences at intermediate timesteps, considering the expected future reward when the sequence is completed. In order to do so, we perform $N$-time Monte Carlo search with the canonical rollout policy $G_\theta$ represented as
\begin{equation}
\text{MC}^{G_\theta}(Y_{1:t};N) = \{Y^1_{1:T}, ..., Y^N_{1:T}\}
\end{equation}

where $Y^n_{1:t} = Y_{1:t}$ and $Y^n_{t+1:T}$ is stochastically sampled via the policy $G_\theta$. Now $Q(s,a)$ becomes

\begin{equation}
Q(Y_{1:t-1}, y_t) = 
\begin{cases}
    \frac{1}{N} \underset{n=1..N}{\sum} R(Y^n_{1:T}), \text{with} \\ Y^n_{1:T} \in \text{MC}^{G_\theta}(Y_{1:t}; N), & \text{if $t < T$}.\\
    R(Y_{1:T}), & \text{if $t = T$}.
  \end{cases}
\end{equation}

An unbiased estimation of the gradient of $J(\theta)$ can be derived as

\begin{multline}
\nabla_\theta J(\theta) \simeq \frac{1}{T} \sum_{t = 1,...,T} \mathbb{E}_{y_t \sim G_{\theta}(y_t | Y_{1:t-1})} [  \\
     \nabla_\theta \log G_{\theta}(y_t | Y_{1:t-1}) \cdot Q(Y_{1:t-1}, y_t)  ]
\end{multline}

Finally in SeqGAN the reward function is provided by $D_\phi$. 

\section{ORGAN}

\begin{figure}[h!]
\includegraphics[width=\columnwidth]{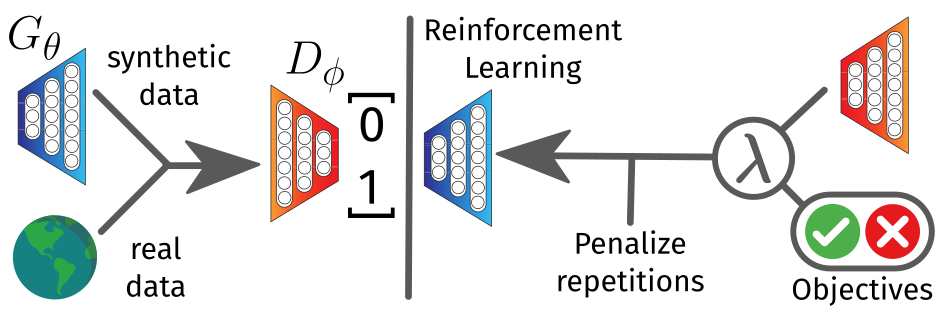}
\caption[Schema for ORGAN]{Schema for ORGAN. \textit{Left}: $D$ is trained as a classifier receiving as input a mix of real data and generated data by $G$. \textit{Right}: $G$ is trained
by RL where the reward is a combination of $D$ and the objectives, and is passed back to the policy function via Monte Carlo sampling. We penalize non-unique sequences.
\label{fig:Schema_ORGAN}}
\end{figure}

Figure \ref{fig:Schema_ORGAN} illustrates the main idea of ORGAN.
To take into account domain-specific desired objectives $O_i$, we extend the reward function for a particular sequence $Y_{1:t}$ to a linear combination of $D_\phi$ and $O_i$, parametrized by $\lambda$:

\begin{equation}
R(Y_{1:T}) = \lambda \cdot D_\phi(Y_{1:T}) + (1 - \lambda) \cdot O_i(Y_{1:T})
\end{equation}

If $\lambda = 0$ the model ignores $D$ and becomes a "naive" RL algorithm, whereas if $\lambda = 1$ it is simply a SeqGAN model. It should be noted that, if chosen, the objective function can vary based on the current iteration of adversarial training, leading to alternating rewards between several objectives and the discriminator.

An additional mechanism to prevent mode collapse is to penalize non-unique sequences by dividing the reward of a repeated sequence by it's the number of copies. The more a sequence gets repeated, the more it will have diminishing rewards. Alternatively, domain-specific similarity metrics could be used to penalize.

To improve the stability of learning, and avoid of problems of GAN convergence like "perfect discriminator", we also implemented the Wasserstein-1 $W$ distance, also known as earth mover's distance, for $D_\phi$ \cite{Arjovsky2017a}. Although the computation of this distance is intractable due to an infimum, it can be transformed via the Kantorovich-Rubinstein duality:

$$
W(p_{data},p_{G}) =\frac{1}{K} \sup_{\|D \| \ge K}  \mathbb{E}_{Y \sim p_{\text{data}}}[D(Y)] - \mathbb{E}_{Y \sim p_{G}}[D(Y)]
$$

Under $W$, $D$ is no longer meant to classify data samples, but now trained and converged to learn $\phi$ such that $D_\phi$ is K-Lipschitz continuous and used to compute the Wasserstein distance. Intuitively the cost of moving the generated distribution to the data. In this context, $D$ can now be considered as a critic in an actor-critic setting. 

\subsection{Implementation Details}

$G_\theta$ is a RNN with LSTM cells, while $D_\phi$ is Convolutional Neural Network (CNN) designed specifically for text classification tasks \cite{CNN}.

To avoid over-fitting with the CNN, we optimized its architecture on classification task between different datasets for each experiment. In the molecule generation task, we utilized a set of drug-like and nondrug-like molecules from the ZINC database \cite{ZINC}. In the music task, we discriminated between a set of folk and videogame tunes scraped from the internet. We utilize a dropout layer at $75\%$ and also $L2$ regularization on the network weights. All the gradient descent steps are done using the Adam algorithm \cite{Adam}.

Molecular metrics are implemented using the RDKit chemoinformatics package \cite{Landrum}. Music metrics employ the MIDI frequencies. The code for ORGAN, including metrics for each experiment, can be found at \url{http://github.com/gablg1/ORGAN}\footnote{Repo soon to be updated (May'18)}.

\section{Experimental results}

In this section, we will test the performance of ORGAN in two scenarios: the generation of molecules encoded as text sequences and musical melodies. Our objective is to show that ORGAN can generate samples that fulfill some desired objectives while promoting diversity. For purposes of interpretation, the range of each objective has been mapped to  $[0,1]$ range, where $0$ corresponds to an undesirable property and $1$ to a very desirable property. Each generator model was pre-trained for 250 epochs using MLE, and the discriminator was trained for 10 epochs.

To measure diversity we use domain-specific measures. In both fields, there are multiple ways of quantifying the notion of diversity so we tried utilizing more widely used metrics.

We compare ORGAN and the Wasserstein variant ($W$) with three other methods of training RNNs: SeqGAN, Naive RL, and Maximum Likelihood Estimation (MLE). Unless specified, $\lambda$ is assumed to be 0.5. All training methods involve a pre-training step of 250 epochs of MLE for $G_\theta$, and 10 epochs for $D_\phi$. The MLE baseline simply stops right after pre-training, while the other methods proceed to further train the model using the different approaches, up to 100 epochs.

For each dataset, we first build a dictionary mapping the vocabulary - the set of all characters present in the dataset - to integers. The dataset is then preprocessed by transforming each sequence into a fixed sized integer sequence of length $N$ where $N$ is the maximum length of a string present in the dataset (in the case of molecules, along with around $10\%$ more characters to increase flexibility and allow generation of larger samples of data). Every string with a length smaller than $N$ is padded with ``\_" characters. Thus the input to our model becomes a list of fixed sized integer sequences.

\subsection{Experiment: Molecules}

Here we test the effectiveness of ORGAN for generating molecules with desirable properties in a pharmaceutical context of drug discovery.

Molecules can be encoded as text sequences by using the SMILES representation \cite{SMILES} of a molecule. This representation encodes the topological information of a molecule based on common chemical bonding rules. For example, the 6-carbon ringed molecule benzene can be encoded as 'C1=CC=CC=C1'. Each C represents a carbon atom, the '=' symbolizes a double bond and '1' the start and closing of a cycle/ring, hydrogen atoms can be deduced via simple rules.

The SMILES representation has predefined grammar rules, and as such, it is possible to have invalid expressions that cannot be decoded back to a valid molecule. Therefore desired property on a generative algorithm is to have a high percentage of valid expression. Invalid expressions get penalized. Additionally, we also penalize the generation of duplicate molecules. 

Recent generative models (\cite{Gomez-Bombarelli2016a},\cite{Kusner2017a}) have reported valid expression rates between $4\%$ up to $80\%$. It should be noted that there are common uninteresting ways to generate valid expressions by alternating "C" and "O" characters such as 'CCCCCCCC' and 'COCCCCOC', the combinatorial possibilities of such permutations is already huge. 

For training, we utilized a random subset of 5k molecules from the set of 134 thousand stable small molecules \cite{134k}. This is a subset of all molecules with up to nine heavy atoms (CONF) out of the GDB-17 universe of 166 billion organic molecules \cite{134k}. The maximum sequence length is 51 and the alphabet size is 43.

When choosing objectives we picked qualities that are normally desired for small molecule drug discovery:

\begin{description}
\item[Solubility:] a property that measures how likely a molecule is able to mix with water, also known as the water-octanol partition coefficient (LogP). Computed via RDKit's Crippen function \cite{Landrum}.
\item[Synthetizability:] estimates how hard (0) or how easy (1) it is to synthesize a given molecule \cite{Ertl2009}.
\item[Druglikeness:] how likely a molecule is a viable candidate for a drug, an estimate that captures the abstract notion of aesthetics in medicinal chemistry \cite{Bickerton2012a}. This property is correlated to the previous two metrics.
\end{description}

To estimate the diversity of our generated samples we can utilize the notion of molecular similarity to construct a measure of how similar or dissimilar a molecule is with respect to a dataset. This measure is based on molecular fingerprints and their Jaccard distance \cite{Sanchez-Lengeling2017}. More concretely, Diversity measures the average similarity of a molecule with respect to a set, in this case, a random subset of molecules from the training set.  
A value of 1 would indicate the molecule is likely to be considered a diverse member of this set, 0 would indicate it has many repeated sub-structures with respect to the set. 

\begin{table*}[!htbp]
\centering
\begin{tabular}{l l c c  r r r r r r}
\toprule
         Objective & Algorithm & Validity (\%) & Diversity & \multicolumn{2}{c}{Druglikeliness}  & \multicolumn{2}{c}{Synthesizability}  &   \multicolumn{2}{c}{Solubility} \\
\midrule
                   &    MLE    &  75.9 &      0.64 &      0.48 &(0\%) &        0.23 & (0\%) &    0.30 &(0\%) \\
                   &    SeqGAN &         80.3 &      0.61 &      0.49 &(2\%) &        0.25 & (6\%) &    0.31  &(3\%) \\ \midrule
    Druglikeliness &     ORGAN &         88.2 &      0.55 &      \graycell 0.52 &\graycell(8\%) &       0.32  &(38\%) &   0.35  &(18\%) \\
     &    OR(W)GAN &          85.0 &      \textbf{0.95} &     \graycell\textbf{0.60} &\graycell\textbf{(25\%)} &      0.54  &(130\%) &   0.47  &(57\%) \\
    &        Naive RL &         97.1 &      0.8  &     \graycell 0.57  &\graycell (19\%) &      0.53  &(126\%) &   0.50  &(67\%) \\\midrule
        Synthesizability &     ORGAN &         \textbf{96.5} &      \textbf{0.92} &      0.51 &(6\%) &     \graycell \textbf{0.83}  & \graycell\textbf{(255\%)} &   0.45  &(52\%) \\
   &    OR(W)GAN &         \textbf{97.6} &     \textbf{ 1.00} &    0.20 &(-59\%) &  \graycell    0.75  & \graycell (223\%) &  0.84  &(184\%) \\
   &    Naive RL &       \textbf{97.7} &      \textbf{0.96} &      0.52 &(8\%) &   \graycell  \textbf{ 0.83}  & \graycell \textbf{(256\%) }&   0.46  &(54\%) \\\midrule
           Solubility &     ORGAN &         \textbf{94.7} &      0.76 &      0.50 &(4\%) &      0.63  &(171\%) &    \graycell 0.55 & \graycell (85\%) \\
        &    OR(W)GAN &         94.1 &     \textbf{ 0.90} &    0.42 &(-12\%) &      0.66  &(185\%) &  \graycell 0.54  & \graycell(81\%) \\
        &        Naive RL &         92.7 &      0.75 &      0.49 &(3\%) &      0.70  & (200\%) &  \graycell \textbf{0.78}  & \graycell \textbf{(162 \%)} \\ \midrule
     All/Alternated  &        ORGAN &         96.1 &      92.3 &  \graycell    0.52 & \graycell (9\%) &    \graycell  0.71  & \graycell (206\%) &  \graycell 0.53  & \graycell (79\%) \\  
\bottomrule
\end{tabular}

 \caption{Evaluation of metrics, on several generative algorithms and optimized for different objectives for molecules. Reported values are mean values of valid generated molecules. The percentage of improvement over the \textit{MLE} baseline is reported in parenthesis. Values shown in bold indicate significant improvement. Shaded cell indicates direct optimized objectives.}
 \label{table mol}
\end{table*}

Table \ref{table mol} shows quantitative results comparing ORGAN to other methods and three different optimization scenarios. MLE and SeqGAN are able to capture the distribution of properties of the training set with minimal alteration in their metrics. While the metric optimized methods excelled in all metrics above the non-optimized methods, effectively showing that they are able to bias the generation process. The Wasserstein variant of ORGAN also seemed to give better diversity properties.

In our experiments, we also noted that naive RL has different failure scenarios. For instance, this approach excelled particularly in the task of Solubility, this particular task rewards very simple sequences such as for the single atom molecule ``N" or monotonous patterns like ``CCCCCCC" or ``CCOCOCCCC" positively. It seems for the other approaches, the GAN/WGAN setting is enforcing more diversity and so punishes these types of patterns, providing highly soluble molecules with more complex features.

\subsection*{Capacity ceiling}

We did notice a form of capacity ceiling in our generation tasks in two forms. The GAN models tended to generate sequences that had the same average sequence length as the training set (15.42). With RL we did not observe this constraint, either it went quite low with synthesizability (9.4) or high (21.3) with druglikeliness. This might be advantageous or detrimental based, on the setting. Optimizing a property that relates to sequence length, for example, molecular size might change this.

\begin{figure}[!htbp]
    \centering
    \includegraphics[width=\linewidth]{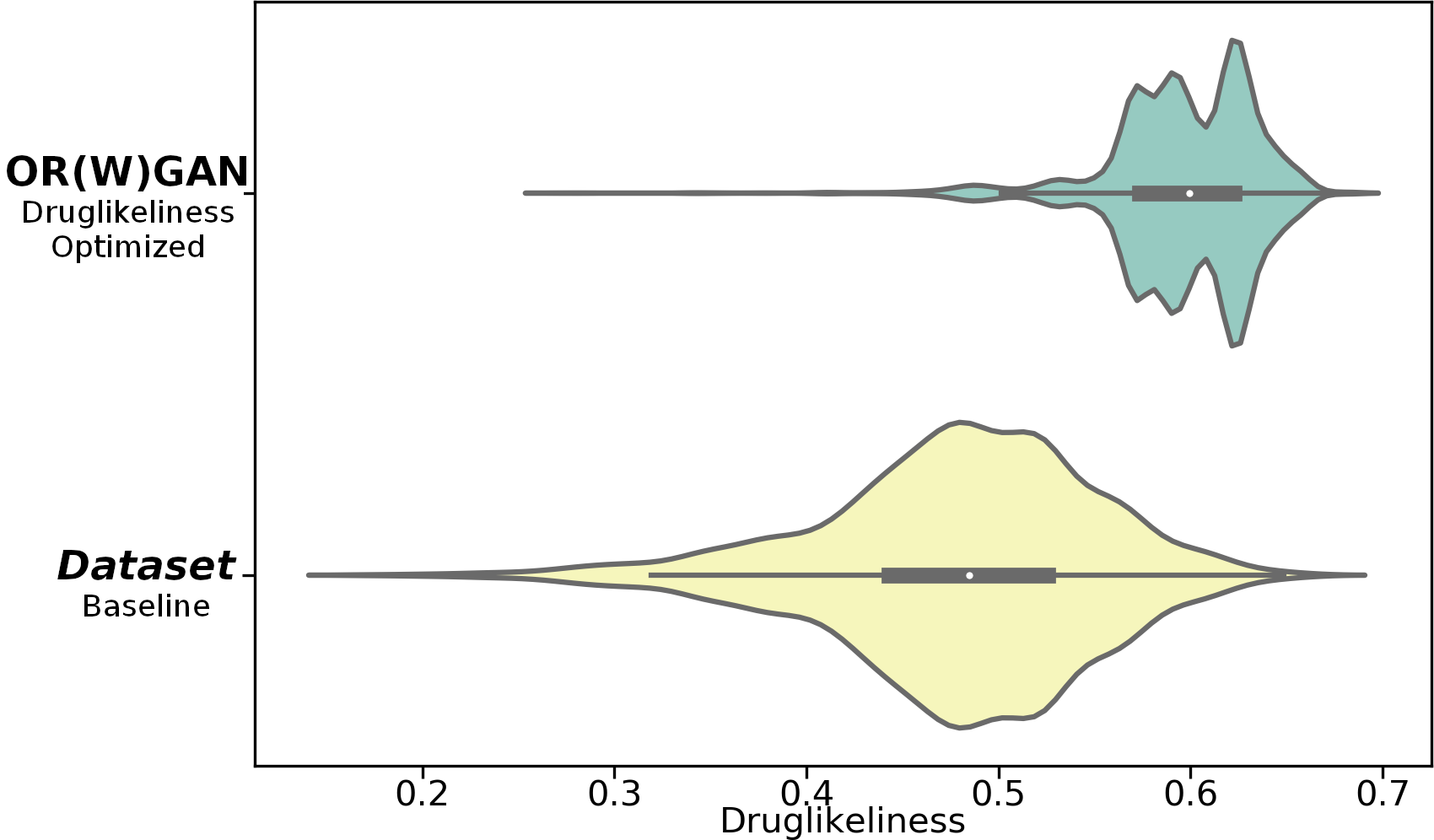}
    \caption{Violinplots of Druglikeliness for molecules from the baseline \textit{Dataset}(n=5000) and optimized OR(W)GAN (n=5440).}
    \label{fig:mol_dist}
\end{figure}

The other ceiling is illustrated in figure \ref{fig:mol_dist}, where the upper limits in Druglikeliness for the data and the best performing approach match. While OR(W)GAN tends to generate more druglike molecules, they do not reach the highest value of 1. This might be property and dataset dependent.

\subsection*{Multi-objective training programs}
\begin{figure}[!htbp]
    \centering
    \includegraphics[width=\linewidth]{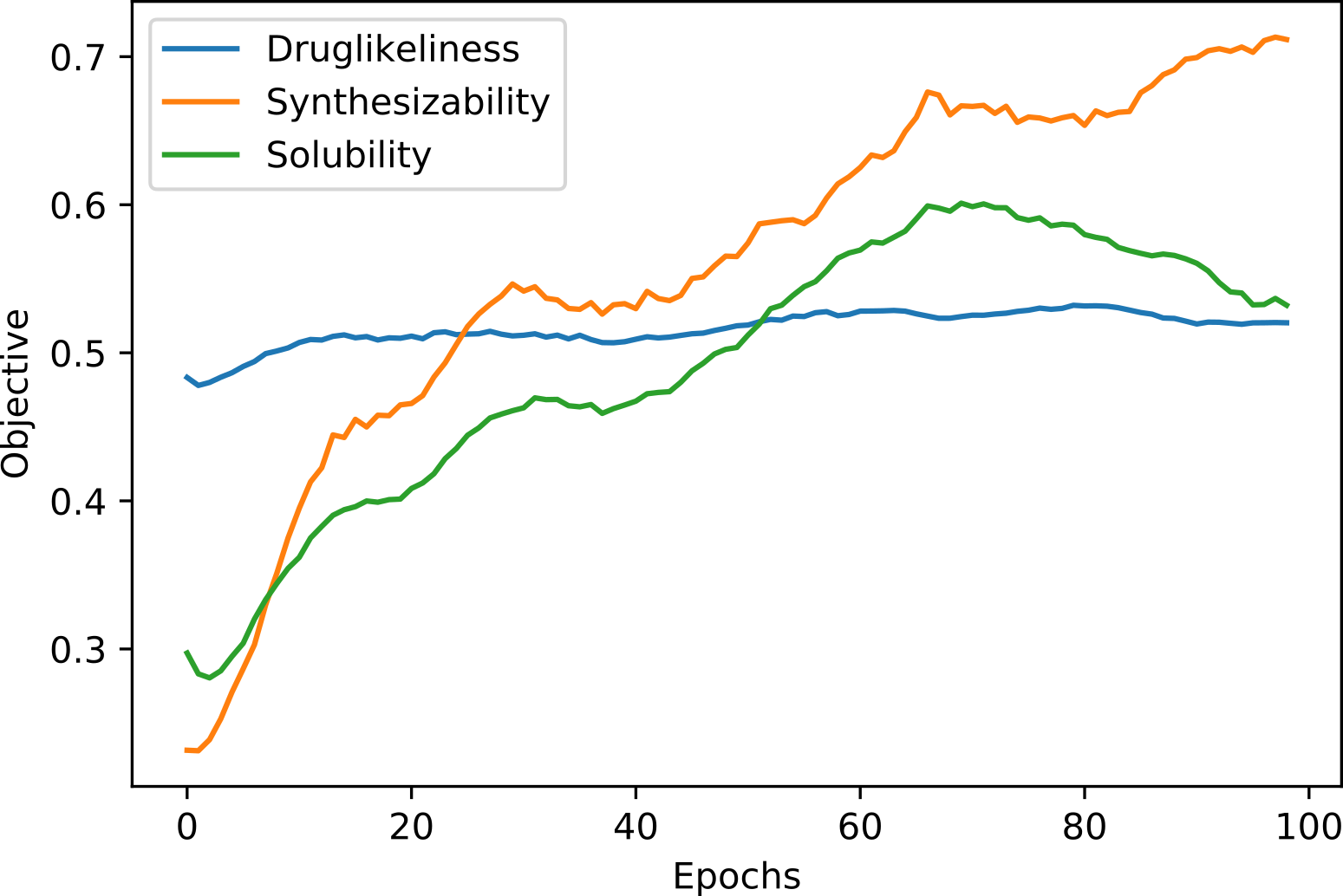}
    \caption{Plots of each objective across the training epochs. Objectives were trained for one epoch, and then switched for another.}
    \label{fig:multi}
\end{figure}

We also experimented with alternating objectives during training. By training for one epoch each objective in rotation until 99 epochs (33 epochs per objective) we arrive to figure \ref{fig:multi}. 

Surprisingly by alternating the objectives, as seen in the last row of table  \ref{table mol}, the gains in each metric are quite high and almost comparable with the best models in each individually trained objective. Although it can also be appreciated in the slight fluctuating behavior of the graphs that there might be limits to the gains that can be achieved. Further work is warranted in this direction.

\subsection{Experiment: Musical melodies}

To further demonstrate the applicability of ORGAN, we extend our study to music sequences. We employ the notation introduced by \cite{jaques2017tuning}, where each token corresponds to a sixteenth of a bar of music. The first two tokens are reserved as 0, which is silent, and 1, which means no event; the other 36 tokens encode three octaves of music, from C3 (MIDI pitch 48) to B5. We use a 1k random sample from the Essen Associative Code (EsAC) folk dataset as processed by \cite{chen2017learning}, where every melody has a duration of 36 tokens (2.25 music bars). We generate songs optimizing two different metrics:

\begin{description}
\item[Tonality.] This measures how many perfect fifths are in the music that is generated. A perfect fifth is defined as a musical interval whose frequencies have a ratio of approximately 3:2. 
These provide what is generally considered pleasant note sequences due to their high consonance.
\item[Ratio of Steps.]  A step is an interval between two consecutive notes of a scale. An interval from C to D, for example, is a step. A skip, on the other hand, is a longer interval. An interval from C to G, for example, is a skip. By maximizing the ratio of steps in our music, we are adhering to the conjunct melodic motion. Our rationale here is that by increasing the number of steps in our songs we make our melodic leaps rarer and more memorable \cite{Bonds2013}.
\end{description}

Moreover, we calculate diversity as the average pairwise edit distance of the generated data \cite{Habrard2008}. We do not attempt to maximize this metric explicitly but we keep track of it to shed light on the trade-off between metric optimization and sample diversity in the ORGAN framework.
Table \ref{table music} shows quantitative results comparing ORGAN to other baseline methods optimizing for three different metrics. ORGAN outperforms SeqGAN and MLE in all of the three metrics. Naive RL achieves a higher score than ORGAN for the Ratio of Steps metric, but it under-performs in terms of diversity, as Naive RL would likely generate very simple rather than diverse songs. In this sense, similar to the molecule case, although the Naive RL ratio of steps score is higher than ORGAN's, the actual generated songs can be deemed much less interesting.

\begin{table*}[!htbp]
\centering
\begin{tabular}{ c c c c c c} \toprule
 Objective & Algorithm & Diversity & Tonality & Ratio of Steps  \\ \toprule
 & MLE  & 0.221 & 0.007 & 0.010 \\
 & SeqGAN & 0.187 & 0.005 & 0.010  \\ \midrule
 \textit{Tonality} & Naive RL  & 0.100 & \graycell 0.478 & 2.9E-05\\
& ORGAN  & \textbf{0.268} &\graycell \textbf{0.372} & 1.78E-04   \\
& OR(W)GAN & \textbf{0.268} & \graycell\textbf{0.177} & 2.4E-04 \\ \midrule
 \textit{Ratio of Steps} & Naive RL & 0.321 & 0.001 &\graycell 0.829  \\
& ORGAN  & \textbf{0.433} &  0.001 & \graycell \textbf{0.632} \\
& OR(W)GAN   & \textbf{0.134} &  5.95E-05 & \graycell \textbf{0.622} \\ \bottomrule
\end{tabular}

 \caption{Evaluation of metrics, on several generative algorithms and optimized for different objectives for melodies. Each measure is averaged over a set of 1000 generated songs. Values shown in bold indicate significant improvement over \textit{MLE} baseline. Shaded cell indicates directly optimized objectives.}
 \label{table music}

\end{table*}

We note that the Ratio of Steps and Tonality have an inverse relationship. This is because two consecutive notes - what qualifies as a step - do not have the frequency ratio of a perfect fifth, which are responsible for increasing tonality. In addition, although the usage of the Wasserstein metric seems to decrease the metrics value, this can be explained as the result of slower training.

\begin{figure}[!htbp]
    \centering
    \includegraphics[width=\linewidth]{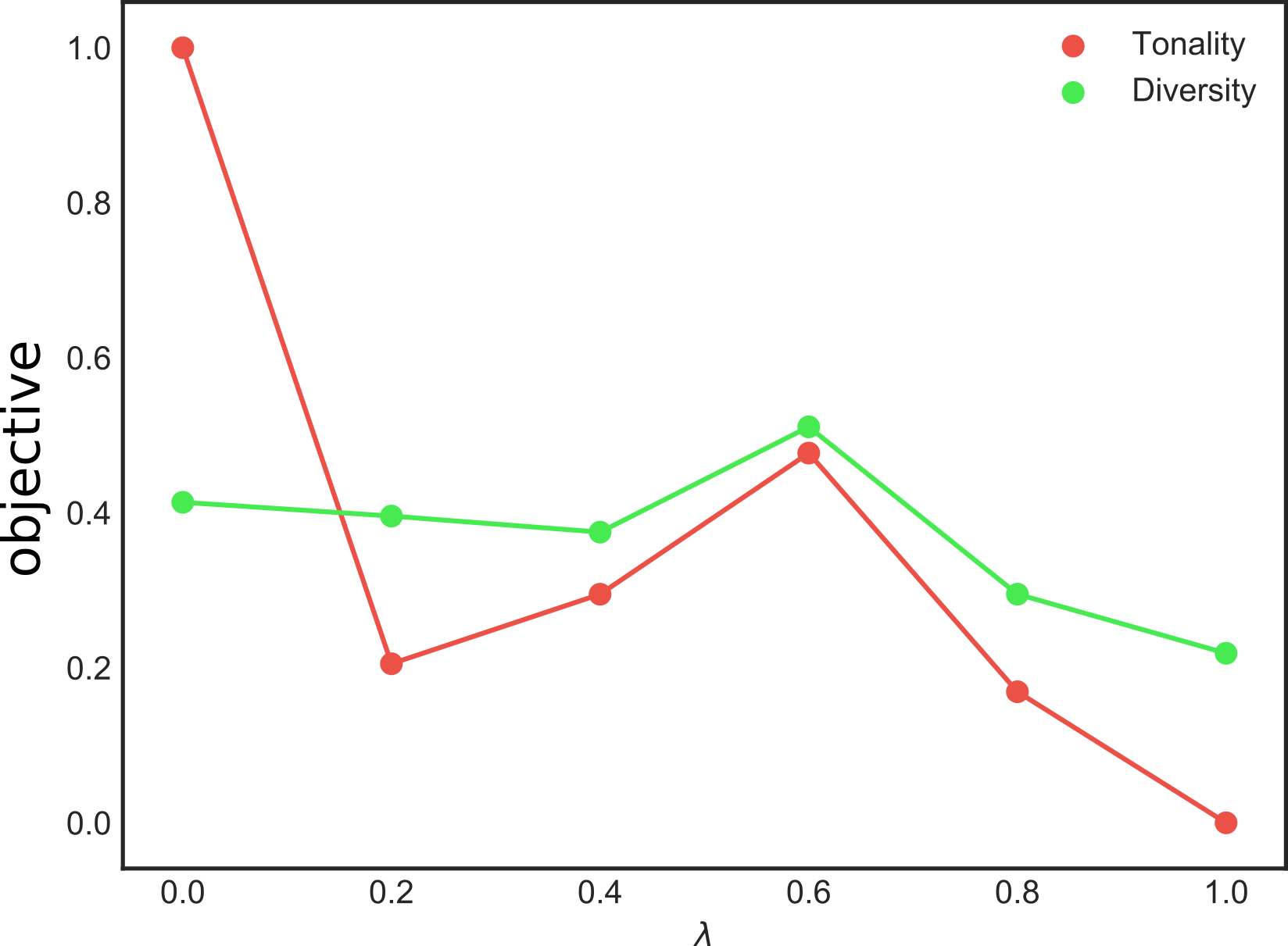}
    \includegraphics[width=\linewidth]{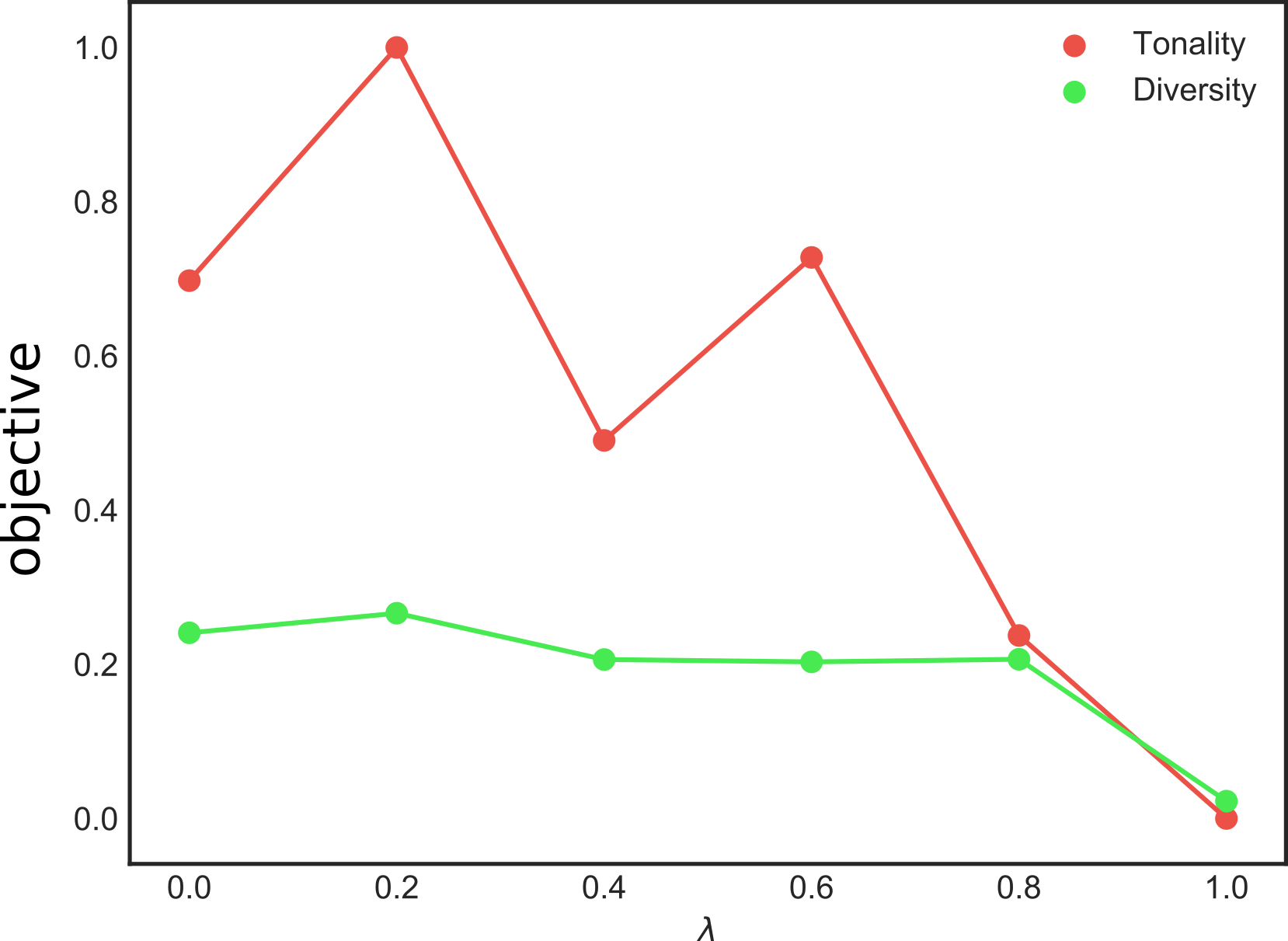}
    \caption{Plots of Diversity and Tonality rewards (the latter re-scaled to the [0, 1] interval) after 80 epochs of training on the music generation task. The upper plot employs the classical GAN loss, while the lower displays a WGAN. The values have been averaged over 1000 samples.}
    \label{fig:lambda_music}
\end{figure}

\subsection*{Effect of $\lambda$}

By tweaking $\lambda$, the ORGAN approach allows one to explore the trade-off between maximizing the desired objective and maintaining likelihood to the data distribution.

Figure \ref{fig:lambda_music} shows the distribution of tonality and diversity sampled from ORGAN and OR(W)GAN for several $\lambda$ values. This showcases that there exists an optimal value for $\lambda$ which maximizes simultaneously the reward and diversity. This value is dependent on the model, dataset and metric, therefore a parameter search would be advantageous to maximize objectives.

\section{Conclusions and future work}

In this work, we have presented ORGAN, a novel framework to optimize an arbitrary object in a sequence generation task. We have built on recent advances in GANs, particularly SeqGAN, and extended them with reinforcement learning to control properties of generated samples.

We have shown that ORGAN can improve certainly desired metrics, achieving better results than recurrent neural networks trained via either MLE or SeqGAN. Even more importantly, data generation can be made subject to a domain-specific reward function while still using the adversarial setting to guarantee the production of non-repetitious samples. Moreover, ORGAN possesses a natural advantage as a black box compared to similar objective-optimization models, since it is not necessary to introduce multiple domain-specific penalties to the reward function: many times a simple objective "hint" will suffice.

As evidenced with the experiments, the RL component is the one of the major drivers for the property optimization and promotion of diversity. Values tended to be higher when RL was present in the architecture of the models.

Future work should investigate how the choice of heuristic can affect the performance of the model. There are also other formulations of GANs for discrete sequences \cite{Che2017},\cite{Hjelm2017} that could be extended with a RL component in order to fine-tune the generation processes.

One area of improvement as seen from figure \ref{fig:mol_dist} is to push the boundaries of the datasets in certain properties. In some domains, outliers might be more valuable such as in the case of drug and materials discovery.

Finally, forthcoming research should extend ORGANs to work with non-sequential data, such as images or audio. This requires framing the GAN setup as a reinforcement learning problem in order to add an arbitrary (not necessarily differentiable) objective function. We believe this extension to be quite promising since real-valued GANs are currently better understood than sequence data GANs.


\bibliographystyle{named}
\bibliography{references}

\end{document}